\documentclass[letterpaper]{article} %
\usepackage{aaai25}  %
\usepackage{times}  %
\usepackage{helvet}  %
\usepackage{courier}  %
\usepackage[hyphens]{url}  %
\usepackage{graphicx} %
\usepackage{natbib}  %
\usepackage{caption} %
\usepackage{amsmath}
\usepackage{amssymb}
\usepackage{mathtools}
\usepackage{amsthm}

\theoremstyle{plain}
\newtheorem{theorem}{Theorem}[section]

\newtheorem{lemma}[theorem]{Lemma}

\theoremstyle{definition}

\theoremstyle{remark}

\usepackage{multicol,lipsum,environ}
\usepackage{thmtools,thm-restate}

\usepackage{algorithm}
\usepackage{algpseudocode}

\usepackage[capitalize,noabbrev]{cleveref}
\newcommand{\suchthat}{\;\ifnum\currentgrouptype=16 \middle\fi|\;}

\usepackage[textsize=tiny]{todonotes}

\newcommand{\E}{\mathbb{E}}
\newcommand{\calA}{\mathcal{A}}

\newcommand{\calX}{\mathcal{X}}

\newcommand{\hatP}{\hat{P}}
\newcommand{\hatQ}{\hat{Q}}
\newcommand{\hatR}{\hat{R}}
\newcommand{\hatV}{\hat{V}}

\newcommand{\supscript}[2]{{#1}^{(#2)}}

\DeclareMathOperator*{\argmax}{arg\,max}

\usepackage{newfloat}
\usepackage{listings}
\DeclareCaptionStyle{ruled}{labelfont=normalfont,labelsep=colon,strut=off} %
\floatstyle{ruled}
\newfloat{listing}{tb}{lst}{}
\floatname{listing}{Listing}
\title{Selective Uncertainty Propagation in Offline RL}
\author{
    Sanath Kumar Krishnamurthy\textsuperscript{\rm 1},
    Tanmay Gangwani\textsuperscript{\rm 2},
    Sumeet Katariya\textsuperscript{\rm 1}, 
    Branislav Kveton\textsuperscript{\rm 3}, 
    Shrey Modi\textsuperscript{\rm 4}, 
    Anshuka Rangi\textsuperscript{\rm 2}
}
\affiliations{
    \textsuperscript{\rm 1}Meta\\ 
    \textsuperscript{\rm 2}Amazon\\ 
    \textsuperscript{\rm 3}Adobe\\ 
    \textsuperscript{\rm 4}Indian Institute of Technology, Bombay\\
    sanathsk@meta.com
}

\begin{document}

\maketitle

\begin{abstract}
We consider the finite-horizon offline reinforcement learning (RL) setting, and are motivated by the challenge of learning the policy at any step $h$ in dynamic programming (DP) algorithms. To learn this, it is sufficient to evaluate the treatment effect of deviating from the behavioral policy at step $h$ after having optimized the policy for all future steps. Since the policy at any step can affect next-state distributions, the related distributional shift challenges can make this problem far more statistically hard than estimating such treatment effects in the stochastic contextual bandit setting. However, the hardness of many real-world RL instances lies between the two regimes. We develop a flexible and general method called selective uncertainty propagation for confidence interval construction that adapts to the hardness of the associated distribution shift challenges. We show benefits of our approach on toy environments and demonstrate the benefits of these techniques for offline policy learning.
\end{abstract}

\section{Introduction}\label{sec:intro}

We study the finite-horizon offline reinforcement learning (RL) problem, focusing on algorithms that adapt to instance hardness. At a high-level, we study algorithms that provide better guarantees for contextual bandit (CB) like instances while being able to plan in more dynamic RL-like instances.

Our work is motivated by real-world RL problems, such as user interaction with an e-commerce search engine (recommendation system). Here, the state can be a user query, and the action is the product recommendation from the engine. When the user wants to buy a particular product, the user often only enters a single product query unrelated to the previous one; thus resembling a sequence of CB problems. On the other hand, when the user explores products, the exploration queries are related through the user's intent, and the recommendation system may want to steer the user toward the ideal product. Hence, this resembles the RL setting. This indicates the need to develop unified solutions that integrate CB and RL techniques -- adapting to instance hardness. We now introduce the CB and RL frameworks in more detail.

Stochastic contextual bandits (CBs) \citep{langford08epochgreedy,li10contextual} and finite-horizon \emph{reinforcement learning (RL)} \citep{sutton88learning,williams92simple,sutton98reinforcement} are two fundamental frameworks for decision-making under uncertainty. In stochastic CBs, the environment samples the context and corresponding rewards (for each action) from a fixed but unknown distribution; the agent then observes the context and learns to select the most rewarding action conditioned on the context. 

Finite-horizon RL is a generalization of CBs where contexts become \emph{states} and a sequence of decisions are to be made over $H$ steps. Similar to the CB problem, at each step, 
the agent observes the current state, selects an action conditioned on the current state, and receives a reward sampled by the environment from a corresponding conditional distribution. However, unlike the CB problem, while the initial state is sampled from a fixed but unknown distribution, the next state at any step depends on the current state and the agent's action. Hence, the agent can plan to attain high cumulative reward by learning to reach high-value future states.

Unfortunately, the fact that actions can influence future states implies that the agent needs to learn under state-distribution shifts making the RL setting much more statistically harder than CBs in the worst case. For example, \citep{foster2021offline} show that the worst-case sample complexity to learn a non-trivial offline RL policy is either polynomial in the state space size or exponential in other parameters.\footnote{\citep{foster2021offline} consider the discounted infinite horizon offline RL formulation. However, one should expect similar lower bounds for the finite horizon offline RL formulation.} On the other hand, if actions do not influence next-state distributions at any step, the RL instance would be equivalent to solving $H$ stochastic CB instances. On such instances, offline bandit algorithms \citep{foster2019orthogonal} would enjoy a polynomial sample complexity for policy learning with no dependence on state space size. Hence, for such instances, state-of-the-art offline RL algorithms such as pessimistic value function optimization \citep{jin2021pessimism} may be unnecessarily conservative.

We formalize this dichotomy and show that the statistical hardness of offline RL instances can be captured by the size of actions' impact on the next state's distribution. To show this, we consider the high-level structure of dynamic programming (DP) algorithms for offline RL \citep[e.g.][]{jin2021pessimism}. DP algorithms construct a policy iteratively starting from the policy for the final step and ending by constructing the policy for the first step. At any step $h$, DP algorithms can be viewed to select the policy at step $h$ that maximizes the treatment effect of deviating from the behavioral policy at step $h$ after having optimized the policy for all future steps. The goal of this paper is to estimate and construct good confidence intervals for this treatment effect at step $h$. 

Our primary focus is on confidence interval (CI) construction, which is motivated by the fact that many successful offline RL algorithms learn a policy that maximizes the lower bound of constructed CIs \citep{jin2021pessimism}. To account for estimation errors from future steps, standard methods for CI construction at any step propagate uncertainty from future steps to the current step $h$. This paper seeks to construct better CIs that adapt to instance hardness by selectively propagating uncertainty. 
In cases where all actions have zero estimated impact on next-state distributions, our procedure does not propagate any uncertainty from later steps and still constructs valid CIs for the treatment effect of deviating from the behavioral policy at step $h$ after having optimized the policy for all future steps. It treats the instance like a CB problem -- hence enjoying a polynomial sample complexity with no dependence on state space size for treatment effect estimation. For more dynamic instances, our procedure must unavoidably propagate more uncertainty from future steps in order to continue constructing valid CIs. In this way, we adapt to the hardness of the instance for CI construction at any step. We also show the benefits of this approach for offline policy learning by proposing an algorithm that optimizes our constructed CIs. Simple simulations further support our claim.

\textbf{Related Work:} Both bandits and RL have been studied extensively \citep{lattimore19bandit,sutton98reinforcement,foster2023foundations}. In bandits, the focus has been on achieving higher statistical efficiency by using the reward distribution of actions \citep{garivier11klucb}, prior distribution of model parameters \citep{thompson33likelihood,agrawal12analysis,chapelle11empirical,russo18tutorial}, parametric structure \citep{dani08stochastic,abbasi-yadkori11improved,agrawal13thompson}, or agnostic methods \citep{agarwal14taming}. In RL, the focus has been on different means of learning to plan for longer horizons, such as the value function \citep{sutton88learning}, policy \citep{williams92simple}, or their combination \citep{sutton00policy}. Just as in our work, causal inference insights have helped improve the statistical efficiency of both CB and RL algorithms \citep{krishnamurthy2018semiparametric,carranza2023flexible,syrgkanis2023post}. However, bridging the gap between bandits and RL is an exciting and relatively under-explored research direction. One way to define this gap is to argue that in bandit-like environments, the state never changes once initially sampled. These bandit-like environments can be viewed as a special case of the situation where actions do not impact next-state distributions. With bridging this gap as one motivation, \citep{zanette2019tighter,yin2021towards} have used variance-dependent Bernstein bounds to limit uncertainty propagation when there is a lack of next-step value function heterogeneity. Another approach is to define this gap in a binary fashion. Either there is no impact of actions on next state distributions, or we are in a dynamic MDP environment. In an online setting, \citep{zhang2022bayesian} develop hypothesis tests to differentiate between the two situations and then select the most appropriate exploration algorithm. While their higher-level framing is similar to ours and their approach is novel, their approach cannot outperform existing RL algorithms in MDP environments. By interpolating between the two regimes, we hope to outperform bandit and existing RL algorithms that either forgo planning or are too conservative in accounting for actions' impact on next-state distributions. 

\section{Preliminaries}
\label{sec:preliminaries}

\textbf{Setting:} We consider an episodic Markov Decision Process (MDP) setting with state space $\calX$, action space $\calA$, horizon $H$, and transition kernel $P=(\supscript{P}{h})_{h=1}^H$. At every episode, the environment samples a starting state $\supscript{x}{1}$ and a set of realized rewards $r=(\supscript{r}{h})_{h=1}^H$ from a fixed but unknown distribution $D$. Here $\supscript{r}{h}$ is a map from $\calX\times\calA$ to $[0,1]$. For any states $x,x'\in\calX$ and action $a\in\calA$, $\supscript{P}{h}(x'|x,a)$ denotes the probability density of transitioning to state $x'$ conditional on taking action $a$ at state $x$ during step $h$. A trajectory $\tau$ is a sequence of states, actions, and rewards. That is, any trajectory $\tau$ is given by $\tau=(\supscript{x}{h},\supscript{a}{h},\supscript{r}{h}(\supscript{x}{h},\supscript{a}{h}))_{h=1}^H$.

A policy $\pi$ is a sequence of $H$ action sampling kernels $\{\supscript{\pi}{h}\}_{h=1}^H$, where  $\supscript{\pi}{h}(a|x)$ denotes the probability of sampling action $a$ at state $x$ during step $h$ under the policy $\pi$. We let $D(\pi)$ denote the induced distribution over trajectories under the policy $\pi$. For any policy $\pi$, we define the (state-) value function $\supscript{V}{h}_{\pi}:\calX\rightarrow [0,H-h+1]$ at each step $h\in[H]$ such that,
\begin{equation}
\label{eq:define-value-function}
    \supscript{V}{h}_{\pi}(x) = \E_{D(\pi)}\bigg[ \sum_{i=h}^H \supscript{r}{i}(\supscript{x}{i},\supscript{a}{i}) \bigg| \supscript{x}{h} = x  \bigg].
\end{equation}
The value of policy $\pi$ is given by $\E_{D}[\supscript{V}{1}_{\pi}(\supscript{x}{1})]$. We can take expectation over $D$ instead of $D(\pi)$ here since the only random variable in $\supscript{V}{1}_{\pi}(\supscript{x}{1})$ is the initial state $\supscript{x}{1}$ which does not depend on the choice of the policy $\pi$.  %

For any step $h\in[H]$, we let $\supscript{R}{h}$ be a function from $\calX\times\calA$ to $[0,1]$ denoting the expected reward function for step $h$.  That is, $\supscript{R}{h}(x,a)=\E_D[\supscript{r}{h}(x,a)]$. With some abuse of notation, for any $x,x'\in\calX$, we let $\supscript{R}{h}(x,\pi)=\sum_a \supscript{\pi}{h}(a|x)\supscript{R}{h}(x,a)$ and $\supscript{P}{h}(x'|x,\pi)=\sum_a \supscript{\pi}{h}(a|x)\supscript{P}{h}(x'|x,a)$. That is, $\supscript{R}{h}(x,\pi)$ is the expected reward at state $x$ and step $h$ under the policy $\pi$. Similarly, $\supscript{P}{h}(x'|x,\pi)$ is the expected transition probability from $x$ to $x'$ at step $h$ under the policy $\pi$. For any step $h\in[H]$, we also let $\supscript{V_{\text{max}}}{h}$ denote a bound on the maximum value $\supscript{V}{h}_{\pi}(x)$ can take for any state $x$ and policy $\pi$. %

It is also equivalent to define the value functions ($\supscript{V}{h}_{\pi}$) using the iterative definition in \eqref{eq:value-iteration-definition}, where $\supscript{V_{\pi}}{H+1} \equiv 0$. 
\begin{equation}
    \label{eq:value-iteration-definition}
    \begin{aligned}
        &\forall h\in[H], x\in\calX,\\ 
        &\supscript{V_{\pi}}{h}(x) = \supscript{R}{h}(x,\pi) + \int_{x'} \supscript{V_{\pi}}{h+1}(x')\supscript{P}{h}(x'|x,\pi).
    \end{aligned}
\end{equation}
\textbf{Data Collection Process:} In this paper, we focus on the offline setting \citep{levine2020offline} with training data collected under a behavioral policy $\pi_b$. Apart from the policy $\pi_b$, the learner only has access to a dataset $S$ consisting of $T$ trajectories sampled from the distribution $D(\pi_b)$, where $D(\pi_b)$ is the data sampling distribution induced by $\pi_b$. That is, $S=\{\tau_t\}_{t=1}^{T}$, where $\tau_t=(\supscript{x}{h}_t,\supscript{a}{h}_t,\supscript{r}{h}_t(\supscript{x}{h},\supscript{a}{h}))_{h=1}^H \sim D(\pi_b)$. Since the transitions in these trajectories are induced by the behavioral policy, for notational convenience, we let $P_b=(\supscript{P_b}{h})_{h=1}^H$ denote the transition kernel under the policy $\pi_b$. That is, for any $x,x'\in\calX$, $\supscript{P_b}{h}(x'|x)=\supscript{P}{h}(x'|x,\pi_b)$. 

\subsection{Estimand of Interest}
We now turn our attention to defining our target estimand, which refers to the specific quantity we aim to estimate. Consider a fixed policy $\pi$ and suppose we would like to estimate its value. Since estimating the value of the behavioral policy $\pi_b$ is easy (empirical average of total observed reward in each trajectory), we argue that that it is sufficient to estimate $\E_{D}[\supscript{V}{1}_{\pi}(\supscript{x}{1})-\supscript{V}{1}_{\pi_b}(\supscript{x}{1})]$ -- the difference in values between evaluation and behavioral policy. This difference can be further decomposed. For each step $h$, let $\tilde{\pi}_{h} = (\supscript{\pi_b}{1}, \dots, \supscript{\pi_b}{h-1},\supscript{\pi}{h},\dots,\supscript{\pi}{H})$ be the policy that follows the behavioral policy upto step $h-1$ and then follows the evaluation policy. In \eqref{eq:decomposing-TE}, we decompose the difference in policy value between the evaluation and behavioral policy into the sum of differences in policy value between $\tilde{\pi}_{h}$ and $\tilde{\pi}_{h+1}$ for each step $h$.
\begin{equation}
\label{eq:decomposing-TE}
    \begin{aligned}
        & \E_{D}[\supscript{V}{1}_{\pi}(\supscript{x}{1}) - \supscript{V}{1}_{\pi_b}(\supscript{x}{1})]\\ 
        & \stackrel{(i)}{=} \sum_{h=1}^H \E_{D}[\supscript{V}{1}_{\tilde{\pi}_h}(\supscript{x}{1}) - \supscript{V}{1}_{\tilde{\pi}_{h+1}}(\supscript{x}{1})]\\
        & \stackrel{(ii)}{=} \sum_{h=1}^H \E_{D(\pi_b)}[\supscript{V}{h}_{\tilde{\pi}_h}(\supscript{x}{h}) - \supscript{V}{h}_{\tilde{\pi}_{h+1}}(\supscript{x}{h})].
    \end{aligned}
\end{equation}
Here (i) follows from telescoping and (ii) follows from the fact that the policies $\tilde{\pi}_h$ and $\tilde{\pi}_{h+1}$ agree with the behavioral policy for the first $h-1$ steps. \emph{We let $\supscript{\alpha}{h}_{\pi}$, the term corresponding to step $h$ in the above decomposition, be our estimand of interest.} That is, our estimand ($\supscript{\alpha}{h}_{\pi}$) is the difference in value of policies $\tilde{\pi}_{h}$ and $\tilde{\pi}_{h+1}$ -- these policies only differ in the current step $h$, which may cause difference in immediate rewards and may also cause a difference in in next-state distributions (affecting future rewards even if the policies at future steps are the same).
\begin{equation}
\label{eq:estimand-of-interest}
    \begin{aligned}
        & \supscript{\alpha}{h}_{\pi} = \E_{D(\pi_b)}[\supscript{V}{h}_{\tilde{\pi}_h}(\supscript{x}{h}) - \supscript{V}{h}_{\tilde{\pi}_{h+1}}(\supscript{x}{h})] 
    \end{aligned}
\end{equation} 
We now seek to justify $\supscript{\alpha}{h}_{\pi}$ as an important estimand, and start by arguing that it is a reasonable estimand to care about. Note that, given the decomposition in \eqref{eq:decomposing-TE}, estimating and constructing CIs for $\{\supscript{\alpha}{h}_{\pi}\}_{h=1}^H$ allows us to estimate and construct CIs for $\E_{D}[\supscript{V}{1}_{\pi}(\supscript{x}{1})-\supscript{V}{1}_{\pi_b}(\supscript{x}{1})]$ (the difference in policy value between evaluation and behavioral policies) -- and thus allows us to estimate and construct CIs for $\E_{D}[\supscript{V}{1}_{\pi}(\supscript{x}{1})]$ (evaluation policy value). 

Beyond being an effective surrogate for policy evaluation, $\supscript{\alpha}{h}_{\pi}$ is an important quantity to consider in dynamic programming (DP) algorithms. DP algorithms construct the policy for the final step ($\supscript{\pi}{H}$) and iteratively construct policies for earlier steps. At step $h$, the policy at steps $h+1$ to $H$ are already fixed/computed. Hence at this step, one can interpret DP algorithms as attempting to select $\supscript{\pi}{h}$ in order to maximize $\supscript{\alpha}{h}_{\pi}$ -- that is, maximize the treatment effect of deviating from the behavioral policy at step h after having optimized the policy for all future steps. Hence, for any step $h$, $\supscript{\alpha}{h}_{\pi}$ is a helpful estimand to consider for decision-making at step $h$.

Importantly for us, when actions at step $h$ do not affect next state distributions, the problem of choosing a policy at step $h$ can be viewed as a CB problem. Helpfully in this case, unlike policy value, $\supscript{\alpha}{h}_{\pi}$ only depends on immediate rewards and can be estimated via offline stochastic CB techniques. However, when actions at step $h$ do influence next state distributions, RL techniques are necessary for estimating $\supscript{\alpha}{h}_{\pi}$. Hence, beyond being a critical quantity for decision-making at step $h$, it is also a quantity that is amenable to interpolating between CB and RL techniques. Thus, our paper focuses on estimating and constructing tight confidence intervals (CIs) for this estimand ($\supscript{\alpha}{h}_{\pi}$).

\section{Shift Model}
\label{sec:shift-model}

Offline RL is more challenging than offline policy learning in the stochastic CB setting \citep{foster2021offline}. The primary reason for the difference between the two settings is due to state distribution shift induced due to deviating from the behavioral policy. Distribution shift makes any statistical learning theory problem challenging \citep{vergara2012chemical,bobu2018adapting,farshchian2018adversarial}. Hence methods that adapt to instance hardness must rely on some implicit or explicit approach to measure this state-distribution shift. To this end, we model the ``heterogeneous treatment effect" \citep{kunzel2019metalearners,nie2021quasi} of actions on the next-state distribution and refer to this effect as the ``shift model". More precisely, we define the shift model $\Delta=(\supscript{\Delta}{h})_{h=1}^H$ in \eqref{eq:shift-definition}.    
\begin{equation}
    \label{eq:shift-definition}
    \begin{aligned}
    &\forall (x,a), \supscript{\Delta}{h}(\cdot|x,a) = \supscript{P}{h}(\cdot|x,a) - \supscript{P_b}{h}(\cdot|x).
    \end{aligned}
\end{equation} %
Here $\Delta^{(h)}(x'|x, a)$ captures the shift in the probability of transitioning from $x$ to $x'$ due to selecting action $a$ at state $x$ instead of following the behavioral policy. With some abuse of notation, for any $x,x'\in\calX$, we let $\supscript{\Delta}{h}(x'|x,\pi)=\sum_a \supscript{\pi}{h}(a|x)\supscript{\Delta}{h}(x'|x,a)$. That is, $\supscript{\Delta}{h}(x'|x,\pi)$ is the expected shift (w.r.t to $P_b$) in probability of transitioning from $x$ to $x'$ at step $h$ under the policy $\pi$. 
It is worth noting that shifts are bounded. For all $(x,a)$, since the $\Delta^{(h)}(\cdot|x, a)$ is a difference of two state-distributions, we have $\|\Delta^{(h)}(\cdot|x, a)\|_1\leq 2$ from triangle inequality.

We argue that shift helps capture instance hardness for estimating $\supscript{\alpha}{h}_{\pi}$. To see this, we provide a shift-dependent expression for $\supscript{\alpha}{h}_{\pi}$.
\begin{equation}
\label{eq:connecting-estimand-to-shift}
    \begin{aligned}
        \supscript{\alpha}{h}_{\pi} &\stackrel{(i)}{=} \E_{D(\pi_b)}[\supscript{V}{h}_{\tilde{\pi}_h}(\supscript{x}{h}) - \supscript{V}{h}_{\tilde{\pi}_{h+1}}(\supscript{x}{h})] \\
        & \stackrel{(ii)}{=} \E_{D(\pi_b)}[\supscript{R}{h}(\supscript{x}{h},\pi) -  \supscript{R}{h}(\supscript{x}{h},\pi_b)] \\
        & + \E_{D(\pi_b)}\bigg[ \int_{x'} \supscript{V_{\pi}}{h+1}(x')\supscript{\Delta}{h}(x'|\supscript{x}{h},\pi) \bigg].
    \end{aligned}
\end{equation} 
Here (i) follows from \eqref{eq:estimand-of-interest} (definition of $\supscript{\alpha}{h}_{\pi}$); and (ii) follows from \eqref{eq:value-iteration-definition} and \eqref{eq:shift-definition}. Note that, in the final expression of \eqref{eq:connecting-estimand-to-shift}, the first term can be estimated using stochastic CB techniques and the dependence on next-step value function is scaled by the size of this shift. This hints at the possibility of developing methods that interpolate between CB and RL techniques. More formally, in \Cref{sec:selective-uncertainty-propagation-theory}
, we show shift estimates enable us to adapt to the hardness of our setting -- when estimating and constructing CIs for $\supscript{\alpha}{h}_{\pi}$.

\section{Theory: Selective Propagation}
\label{sec:selective-uncertainty-propagation-theory}

In \Cref{sec:preliminaries}, we motivated and defined our estimand $\supscript{\alpha}{h}_{\pi}$ (see \eqref{eq:estimand-of-interest}) -- which is the treatment effect for deviating from the behavioral policy at step $h$ after having already deviated from the behavioral policy for all future steps. We now present an approach to estimate and construct tight valid CIs for $\supscript{\alpha}{h}_{\pi}$ -- with interval size adapting to instance hardness. Here harder instances have a larger next-state distribution shifts when deviating from the behavioral policy. When shifts are smaller, we can rely more on statistically efficient CB methods. However when shifts are larger (instance is more dynamic), we unavoidably must rely more on RL methods that account for worst-case distribution shifts.

Our approach to estimate and construct tight valid CIs for $\supscript{\alpha}{h}_{\pi}$ requires several inputs. These inputs, described in the following subsection, allow us to abstract away existing approaches to tackle well-studied estimation problems in CB and RL settings. In \Cref{sec:combining-inputs}, we describe how to combine these existing tools to achieve guarantees that adapt to instance hardness.

\subsection{Inputs}
\label{sec:inputs}

Our method interpolates between existing tools for CB and RL settings, by leveraging shift estimates. To simplify our analysis and generalize our results, we assume access to these estimates as inputs to our interpolation method. In particular, we take as input: (1) offline CB treatment effect estimate and corresponding CI, (2) optimistic and pessimistic offline RL value function estimates, and (3) shift estimates with average error bounds. As the quality of our inputs improve (potentially as better estimators get developed), the quality of our outputs will correspondingly improve.

We now formally describe these inputs -- requiring all the associated high-probability bounds to hold simultaneously with probability at least $1-\delta_{\text{in}}$. We start by describing the first input, which is based on CB methods.

\textbf{Input 1 (CB estimates)}: This input provides an estimate and CI for $\supscript{\theta}{h}$ (formally defined in \eqref{eq:immediate-TE-estimand}) -- which is the average treatment effect on the immediate reward for deviating from the behavioral policy at step $h$.
\begin{equation}
\label{eq:immediate-TE-estimand}
    \supscript{\theta}{h}_{\pi} = \E_{D(\pi_b)}\Big[\supscript{R}{h}(\supscript{x}{h},\tilde{\pi}_h) - \supscript{R}{h}(\supscript{x}{h},\tilde{\pi}_{h+1})\Big]
\end{equation} 
Since $\supscript{\theta}{h}_{\pi}$ only depends on the immediate reward, well-established offline CB techniques \citep[e.g., ][]{dudik14doubly} can be used to estimate and construct CIs for the difference (in terms of immediate rewards) between these policies. We let $\supscript{\hat{\theta}}{h}_{\pi}$ be our input estimate and let $\supscript{\kappa}{h}_{\pi,\theta}$ be the input CI radius. That is, the confidence interval is given by \eqref{eq:immediate-TE-CI}.
\begin{equation}
\label{eq:immediate-TE-CI}
    |\supscript{\theta}{h}_{\pi} - \supscript{\hat{\theta}}{h}_{\pi}| \leq \supscript{\kappa}{h}_{\pi,\theta}
\end{equation}
When deviating from the behavioral policy at step $h$ has no impact on next-state distributions, the estimate and CI for $\supscript{\theta}{h}_{\pi}$ can be used as the estimate and CI for $\supscript{\alpha}{h}_{\pi}$. However, when there is an impact on next-state distributions, valid estimation and CI construction for $\supscript{\alpha}{h}_{\pi}$ requires us to propagate estimates and uncertainty from future steps to the current step. To enable this propagation, we take estimates for $\supscript{V_{\pi}}{h+1}$ as our second input.

\textbf{Input 2 (RL estimates)}: This input provides pessimistic, standard, and optimistic estimates for $\supscript{V_{\pi}}{h+1}$ -- denoted by $\supscript{\hatV_{\pi, p}}{h+1}$, $\supscript{\hatV_{\pi}}{h+1}$, and $\supscript{\hatV_{\pi, o}}{h+1}$ respectively -- such that the ordering in \eqref{eq:ordering} holds.\footnote{Note that \eqref{eq:ordering} can be enforced during input construction. Recall that $\supscript{V_{\text{max}}}{h+1}$ denotes a bound on the maximum value $\supscript{V}{h+1}_{\pi}(x)$ can take for any state $x$.} Further, with high probability, we require \eqref{eq:next-step-value-function-input-guarantee} holds -- that is, the true value function is bounded by the pessimistic and optimistic value function estimates. 
\begin{align}
    &\forall x,\; 0\leq \supscript{\hatV_{\pi, p}}{h+1}(x) \leq \supscript{\hatV_{\pi}}{h+1}(x) \leq \supscript{\hatV_{\pi, o}}{h+1}(x)\leq \supscript{V}{h+1}_{\text{max}}\label{eq:ordering}\\
    &\forall x,\; \supscript{V_{\pi}}{h+1}(x)\in [\supscript{\hatV_{\pi, p}}{h+1}(x),\supscript{\hatV_{\pi, o}}{h+1}(x)] \label{eq:next-step-value-function-input-guarantee}
\end{align} %
There is a large and growing literature on value function estimation in RL, including optimistic and pessimistic value function estimation that are designed to satisfy \eqref{eq:next-step-value-function-input-guarantee} \citep[e.g.,][]{martin2017count,wang2019optimism,jin2021pessimism}. Thus, we can employ the most cutting-edge methods to construct these next-step value function estimates.

Input 2 gave us estimates for $\supscript{V_{\pi}}{h+1}$ (next-step value), which we may need to propagate to the current step $h$ -- when constructing an estimate and CI for $\supscript{\alpha}{h}_{\pi}$. Since our goal is to interpolate between tight CB guarantees and always valid RL guarantees, unlike traditional RL algorithms, we want to be selective in propagating next-step estimates/uncertainties. Our final input, shift estimates, allows us to only propagate these estimates when required -- enabling our adaptation to instance hardness.

\textbf{Input 3 (Shift estimates)}: This input provides a estimate for $\supscript{{\Delta}}{h}$ (see \eqref{eq:shift-definition}) and an associated error bound -- denoted by $\supscript{\hat{\Delta}}{h}$ and $\supscript{\kappa}{h}_{\pi,\Delta}$ respectively -- such that \eqref{eq:shift-estimate-is-bounded} holds (recall from \Cref{sec:shift-model} that $\supscript{{\Delta}}{h}$ satisfies the same bound). We also require \eqref{eq:estimated-shift-guarantees} holds with high-probability.
\begin{align}
    & \forall (x,a),\; \|\supscript{\hat{\Delta}}{h}(\cdot|x,a) \|_1 \leq 2 \label{eq:shift-estimate-is-bounded} \\
    &\E_{D(\pi_b)} \big\|\supscript{\hat{\Delta}}{h}(\cdot|\supscript{x}{h},\pi) - \supscript{\Delta}{h}(\cdot|\supscript{x}{h},\pi)\big\|_1 \leq \supscript{\kappa}{h}_{\pi,\Delta} \label{eq:estimated-shift-guarantees}
\end{align}
Since the true shift model is a function of the true transition model (see \Cref{sec:shift-model}), shift can be estimated via first estimating transition model and then calculating the treatment effect (due to deviating from the behavioral policy) of transitioning between any pair of states \citep[see ][ for a survey on model-based RL and transition model estimation.]{moerland2023model}. \footnote{As a treatment effect model, shift may also be estimated via heterogeneous treatment effect estimators \citep[e.g.,][]{nie2021quasi, kunzel2019metalearners}.} 

\subsection{Combining Inputs}
\label{sec:combining-inputs}

We now have all our required input estimates, and can state our main result (\Cref{thm:treatment-effect-CI}) on constructing an estimate and CI for $\supscript{\alpha}{h}_{\pi}$ -- in a way that adapts to instance hardness.
\begin{restatable}{theorem}{lemTE}
\label{thm:treatment-effect-CI} 
Suppose we have: (1) CB inputs $(\supscript{\hat{\theta}}{h}_{\pi},\supscript{\kappa}{h}_{\pi,\theta})$; (2) RL inputs $(\supscript{\hatV_{\pi, p}}{h+1},\supscript{\hatV_{\pi}}{h+1},\supscript{\hatV_{\pi, o}}{h+1})$ satisfying \eqref{eq:ordering}; and (3) shift inputs $(\supscript{\hat{\Delta}}{h},\supscript{\kappa}{h}_{\pi,\Delta})$ satisfying \eqref{eq:shift-estimate-is-bounded} -- such that \eqref{eq:immediate-TE-CI}, \eqref{eq:next-step-value-function-input-guarantee}, and \eqref{eq:estimated-shift-guarantees} hold with probability at least $1-\delta_{\text{in}}$.\\ 
Moreover, suppose we have a (holdout) dataset of $T$ trajectories $S=\{\tau_t\}_{t=1}^{T}$ -- sampled from the distribution $D(\pi_b)$ -- that were not used for constructing input estimates.\footnote{Utilizing a holdout set for estimating and constructing CIs for $\supscript{\alpha}{h}_{\pi}$ allows us to treat our input estimates as fixed quantities (independent of the randomness in the sampled holdout dataset).} Our estimate for $\supscript{\alpha}{h}_{\pi}$ is then denoted by $\supscript{\hat{\alpha}}{h}_{\pi}$ and given by \eqref{eq:te-estimator-at-step-h}. %
\begin{equation}
\label{eq:te-estimator-at-step-h}
\begin{aligned}
    &\supscript{\hat{\alpha}}{h}_{\pi} = \supscript{\hat{\theta}}{h}_{\pi}+ \frac{1}{T}\sum_{t=1}^{T}\int_{x'}\supscript{\hatV_{\pi}}{h+1}(x')\supscript{\hat{\Delta}}{h}(x'|\supscript{x}{h}_t,\pi)
\end{aligned}
\end{equation}
Now for some fixed $\delta>0$, with probability at least $1-\delta-\delta_{\text{in}}$, we have the confidence interval in \eqref{eq:simulation-step} holds.
\begin{equation}
\label{eq:simulation-step}
    \begin{aligned}
    &|\supscript{{\alpha}}{h}_{\pi} - \supscript{\hat{\alpha}}{h}_{\pi}|\leq \supscript{L}{h}_{\pi}. 
    \end{aligned}
\end{equation}
Here $\supscript{L}{h}_{\pi}$ is given by \eqref{eq:define-bound}. %
\begin{equation}
\label{eq:define-bound}
    \begin{aligned}
    &\supscript{L}{h}_{\pi}= \supscript{\kappa}{h}_{\pi,\theta} + \supscript{V_{\max}}{h+1}\supscript{\kappa}{h}_{\pi,\Delta} + 6\supscript{V}{h+1}_{\max}\sqrt{\frac{\ln(4/\delta)}{2T}} \\
    &+\frac{1}{T}\sum_{t=1}^{T}\int_{x'} |\E_{D(\pi_b)}[\supscript{\hat{\Delta}}{h}(x'|\supscript{x}{h}_t,\pi)]|  \supscript{\Gamma_{\pi}}{h+1}(x')
    \end{aligned}
\end{equation} 
Here $\supscript{\Gamma_{\pi}}{h+1}$ is the difference between the optimistic and pessimistic estimates -- it captures the uncertainty in the next-step value function estimates. That is, for all $x\in\calX$, $\supscript{\Gamma_{\pi}}{h+1}(x)=\supscript{\hatV_{\pi,o}}{h+1}(x)-\supscript{\hatV_{\pi,p}}{h+1}(x)$.
\end{restatable}
One of the advantages of in-distribution supervised learning is that excess risk bounds only depend on complexity of hypothesis class (and number of training samples), with no dependence on size of feature space \cite[see ][]{shalev2014understanding}. As discussed in \Cref{sec:intro}, the statistical challenges of RL stem from the fact that learning under (state) distribution shifts is hard. For example, without additional assumptions, optimistic/pessimistic value function estimation have an unavoidable polynomial dependency on state-space size \citep{foster2023foundations}. Our goal is to avoid/minimize such dependencies when possible. The key benefit of \Cref{thm:treatment-effect-CI} is that both our estimate $(\supscript{\hat{\alpha}}{h}_{\pi})$ and our CI width $(\supscript{L}{h}_{\pi})$ are ``selective" in propagating/utilizing the RL estimates from input 2 -- allowing us to only suffer from the slower worst-case RL estimation rates on harder instances. To better understand this, let us dig deeper into the terms in our CI width ($\supscript{L}{h}_{\pi}$). 

Note that $\supscript{\kappa}{h}_{\pi,\theta}$ and $\supscript{\kappa}{h}_{\pi,\Delta}$ (see Inputs 1 and 3) bound errors averaged under the behavioral policy state-distribution -- that is, they bound in-distribution average errors. Hence, with appropriate inputs, the first two terms in $\supscript{L}{h}_{\pi}$ shrink quickly with no dependency on state-space size \citep{dudik14doubly,shalev2014understanding}. The third term in $\supscript{L}{h}_{\pi}$, which enjoys a $\mathcal{O}(\sqrt{\log(1/\delta)/T})$ rate, also shrinks quickly to zero and has no dependency on state-space size.

We now only need to discuss the fourth and final term in $\supscript{L}{h}_{\pi}$. Unlike the previous terms which bound in-distribution average errors, this term does depend on per-state (point-wise) errors ($\supscript{\Gamma_{\pi}}{h+1}(x')$). The reason RL algorithms seek to bound per-state errors is because these guarantees do not depend on the state-distribution and are valid under any shift. We now illustrate how this robustness to state-distribution shift comes at a cost of larger error bounds, and argue that it is advantageous to scale these terms down with the estimated shift. First, as a sanity check, we show that this term is finite.
\begin{equation}
\label{eq:sanity-check}
    \begin{aligned}
        &\int_{x'} |\supscript{\hat{\Delta}}{h}(x'|\supscript{x_t}{h},\pi)| \cdot \supscript{\Gamma_{\pi}}{h+1}(x') \\ 
        & \stackrel{(i)}{\leq} \|\supscript{\hat{\Delta}}{h}(\cdot|\supscript{x_t}{h},\pi) \|_1  \|\supscript{\Gamma_{\pi}}{h+1}\|_{\infty} \stackrel{(ii)}{\leq} 2 \|\supscript{\Gamma_{\pi}}{h+1}\|_{\infty}. 
    \end{aligned}
\end{equation} %
where (i) follows from H\"older's inequality, and (ii) follows from \eqref{eq:shift-estimate-is-bounded}. Now that we know this term is finite, we can argue that it shrinks to zero. Since $\supscript{\Gamma_{\pi}}{h+1}(x')$ captures the size of per-state errors for pessimistic/optimistic value function estimates, we can expect this term to converges to zero in the limit with infinite data. The size of $\supscript{\Gamma_{\pi}}{h+1}(x')$ must depend on how often states similar to $x'$ were visited at step $h+1$ in the training data for the RL input. For simplicity, let us consider the scenario when all states are visited uniformly at step $h+1$ under the distribution $D(\pi_b)$. In such a scenario, the frequency at which states similar to $x'$ were visited at step $h+1$ would depend on some measure of the size of the state space $\calX$. This would imply that $\supscript{\Gamma_{\pi}}{h+1}(x')$ shrinks at a rate that depends on some measure of the size of the state space $\calX$. That is, while these terms shrink, they shrink slowly. Hence per-state bounds, while independent of state-distribution, come at a cost of slower statistical rates. As shown in \citep{foster2021offline}, such a dependence of confidence interval width on state-space size is unavoidable in the worst-case. %

The key message of \Cref{thm:treatment-effect-CI} is that we can move beyond this worst-case scenario by scaling these point-wise errors with the estimated shifts $(\supscript{\hat{\Delta}}{h})$. For example, when $\supscript{\hat{\Delta}}{h}\equiv 0$, the fourth term in $\supscript{L_{\pi}}{h}$ is zero, allowing us to recover contextual bandit-style guarantees that are independent of state-space size. It is worth noting that, even when state-space size is not a concern, being selective about error/estimate propagation can improve the resulting interval widths.

\section{Modifying Pessimistic Value Iteration}
\label{sec:SPVI}

Pessimistic value iteration (PVI) \citep{jin2021pessimism} is a popular family of dynamic programming (DP) algorithms for offline RL. PVI can take as input: an estimated reward model $\hatR$, an estimated transition model $\hatP$, and point-wise estimation uncertainty bonus $b=(\supscript{b}{h})_{h=1}^H$. No additional data is required. The DP procedure in PVI iterates over steps $h=H$ to $h=1$. For any step of value function estimation, the bonuses $b=(\supscript{b}{h})_{h=1}^H$ must bound the cumulative error in the estimated reward and transition model inputs. \footnote{In finite-state MDPs, these bonuses can be count based, where $\supscript{b}{h}(x,a)=\beta\cdot\sqrt{1/\max\{1,n_h(x,a)\}}$ and $n_h(x,a)$ is the number of times state $x$ and action $a$ are observed at step $h$ in the data set $S$. Here $\beta$ is an algorithmic parameter. Several papers have extended these ideas to continuous state spaces \citep{bellemare2016unifying,osband2021epistemic}.} Hence at any step $h$, \eqref{eq:pessimistic-Q-value} gives a valid pessimistic Q-value estimate ($\supscript{\hatQ_{\text{p}}}{h}$) -- here $\supscript{\hatV_{p}}{h+1}$ is the pessimistic value function for the constructed policy starting from step $h+1$ (computed in the previous dynamic programming step). At every step $h$, PVI selects the policy that maximizes the pessimistic Q-value function.
\begin{equation}
\label{eq:pessimistic-Q-value}
    \supscript{\hatQ_{\text{p}}}{h}(\cdot,\cdot) = \hatR(\cdot,\cdot) + \int_{x'} \supscript{\hatV_{p}}{h+1}(x')\supscript{\hat{P}}{h}(x'|\cdot,\cdot) -\supscript{b}{h}(\cdot,\cdot)
\end{equation}
Pessimistic Q-value maximization helps PVI avoid model exploitation -- that is, PVI avoids picking policies with inaccurately high estimated values at step $h$ by penalizing uncertainty in the value function estimate. However to do this, as we see in \eqref{eq:pessimistic-Q-value}, PVI propagates all the uncertainty captured in future steps through $\supscript{\hatV_{p}}{h+1}$. Depending on the instance, this may lead to larger than necessary uncertainty propagation for avoiding model exploitation. In particular, based on the results in \Cref{sec:selective-uncertainty-propagation-theory}, uncertainty from later steps does not always need to be fully propagated to estimate lower bounds for the effect of deviating from the behavioral policy at step $h$ (after fixing the policy for all future steps). Since we can view maximizing this treatment effect as the goal of DP algorithms at any step $h$, maximizing the tighter lower bounds from \Cref{sec:selective-uncertainty-propagation-theory} should allow us to do better on easier CB-like instances (while avoiding model exploitation).

In this section, we propose a modification of PVI called selectively pessimistic value iteration (SPVI, complete pseudo-code is available in \Cref{sec:spvi-pseudocode}). The key modification is that, at any step $h$, SPVI maximizes the selectively pessimistic Q-value \eqref{eq:selectively-pessimistic-Q-value-main} -- which is the standard Q-value estimate ($\supscript{\hatQ}{h}$) minus the bonus and the required uncertainty that needs to be propagated. Here $\hat{\Delta}$ is the induced shift model (that is, $\supscript{\hat{\Delta}}{h}(\cdot|\cdot,\cdot)= \supscript{\hat{P}}{h}(\cdot|\cdot,\cdot) - \supscript{\hat{P}}{h}(\cdot|\cdot,\pi_b)$), $\supscript{\hatV_{p}}{h+1}$ and $\supscript{\hatV_{o}}{h+1}$ are the pessimistic and optimistic value function for the constructed policy starting from step $h+1$ (computed in the previous dynamic programming step). 
\begin{equation}
\label{eq:selectively-pessimistic-Q-value-main}
\begin{aligned}
    &\supscript{\hatQ_{\text{sp}}}{h}(\cdot,\cdot) = \supscript{\hatQ}{h}(\cdot,\cdot)-\supscript{b}{h}(\cdot,\cdot)\\
    &- \int_{x'} |\supscript{\hat{\Delta}}{h}(x'|\cdot,\cdot)| \cdot (\supscript{\hatV_{o}}{h+1}(x') - \supscript{\hatV_{p}}{h+1}(x'))
\end{aligned}
\end{equation}
\textbf{Justification:} As we argued in earlier sections, one can view the goal at step $h$ as selecting $\supscript{\pi}{h}$ in order to maximize $\supscript{\alpha}{h}_{\pi}$. We construct a tight lower bound for $\supscript{\alpha}{h}_{\pi}$ and argue that maximizing $\supscript{\hatQ_{\text{sp}}}{h}$ maximizes this bound. Similar to standard pessimistic value estimation, we let the bonus $\supscript{b}{h}(x,a)$ bound the total model estimation errors at $(x,a,h)$. Now from the analysis in \Cref{thm:treatment-effect-CI}, we have \eqref{eq:Q-TE-bound} is a valid lowed bound on $\supscript{\alpha}{h}_{\pi}$.
\begin{equation}
\label{eq:Q-TE-bound}
    \begin{aligned}
        &\supscript{\alpha}{h}_{\pi} = \E_{D(\pi_b)}[\supscript{V}{h}_{\tilde{\pi}_h}(\supscript{x}{h}) - \supscript{V}{h}_{\tilde{\pi}_{h+1}}(\supscript{x}{h})] \\
        & \geq \E_{D(\pi_b)}[\supscript{\hatQ_{\text{sp}}}{h}(\supscript{x}{h},\supscript{\pi}{h}) - \supscript{\hatQ}{h}(\supscript{x}{h},\supscript{\pi_b}{h})]
    \end{aligned}
\end{equation}
Importantly, by maximizing this tight lower bound we can do better on easier CB-like instances (while always avoiding model exploitation by penalizing uncertainty in estimating $\supscript{\alpha}{h}_{\pi}$). This completes our justification of SPVI. The complete pseudo-code is available in \Cref{sec:spvi-pseudocode}.

\section{Simulation}
\label{sec:simulation}

\begin{figure}[h]
\begin{center}
\centerline{
\includegraphics[width=0.4\textwidth]{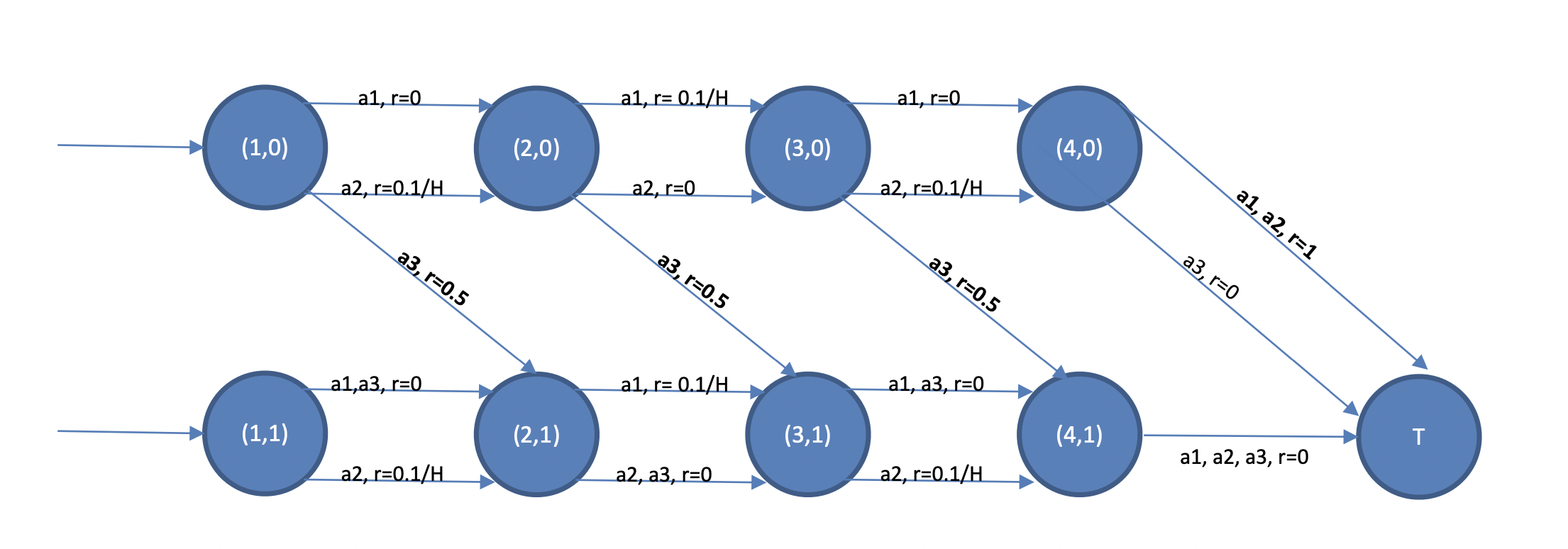}
}
\caption{ChainBandit MDP with horizon/length of 4 (this is an adjustable environment parameter). The environment has two chains, a top chain, and a bottom chain. The environment also has three actions given by $a_1,a_2,a_3$. The top chain states are the most rewarding. 
The agent starts at (1,0). At any state in the bottom chain, all the actions lead to the same transition (which is to move to the next state in the bottom chain) and are essentially bandit states. In the top chain, both $a_1$ and $a_2$ lead to the same transitions (which is to move to the next state in the top chain), and $a_3$ makes the agent move to the next state in the bottom chain. In the top chain, the highest cumulative reward comes from never taking action $a_3$; however, the highest immediate reward comes from selecting the action $a_3$ (which makes planning beneficial in this environment). Note that at every state, action $a_3$ is a sub-optimal action.  }
\label{fig:chainbanditenv}
\end{center}
\vskip -0.25in
\end{figure}

To illustrate our insights, we consider a simple toy environment called ``ChainBandit". As described in \Cref{fig:chainbanditenv}, this environment is constructed to have both dynamic (some actions result in a different next-state distribution) and bandit-like (non-dynamic) elements. Ensuring that while planning and some estimate/uncertainty propagation is necessary, complete uncertainty propagation can be unnecessary to evaluate policies of interest. Throughout this section, we consider ChainBandit with a chain length of 3 and consider the following behavioral policy ($\pi_b$) -- at every state and step, $\pi_b$ selects action $a_3$ with probability $0.8$ and selects the other two actions with probability $0.1$ respectively. From the data collected, we evaluate standard and selective uncertainty propagation for tasks of: (1) estimating upper/lower bounds for $\supscript{\alpha}{h}_{\pi}$; and (2) offline policy learning.

Since uncertainty propagation is the focus of this simulation, in order to make a fair comparison, both standard and selective propagation: use the same tabular approach to estimate a (step independent) reward/transition model; and use the same (step independent) count-based bonuses to account for model estimation errors.\footnote{The model estimates and bonuses are not step dependent since the reward/transition functions in Chain Bandit are the same for all steps. Further a tabular approach to reward (transition) estimation simply indicates using the average reward (average one-hot next-state vector) observed at any state-action pair as its reward (transition) estimate.} Here bonus is given by $b(x,a)=\sqrt{\ln(|\calX|*|\calA|*H/\delta)/n(x,a)}$ -- where $n(x,a)$ is the number of times action $a$ was taken at state $x$, and confidence parameter $\delta = 0.05$. 

\begin{figure}[ht]
\begin{center}
\centerline{
\includegraphics[width=0.4\textwidth]{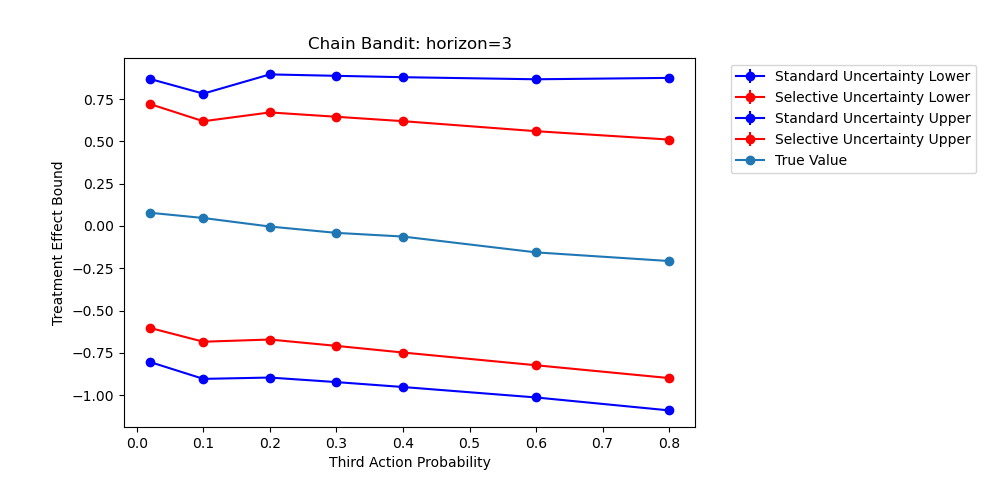}
}
\caption{We plot CIs for $\supscript{\alpha}{2}_{\pi}$ while varying the evaluation policy. These evaluation policies are parameterized by $\lambda\in[0,1]$. For all states/steps, the probability of selecting $a_1, a_2$ and $a_3$ are $(1-\lambda)/2,(1-\lambda)/2,$ and $\lambda$ respectively. Note that the evaluation policy is the same as the behavioral policy for $\lambda=0.8$. The number of training episodes is 10000, and the plots are averaged over 10 runs.}
\label{fig:chainbanditenv-CI}
\end{center}
\vskip -0.25in
\end{figure}

For the step $h$ and policy $\pi$ of interest, standard CIs for $\supscript{\alpha}{h}_{\pi}$ are constructed using pessimistic/optimistic value estimates at step $h$ for policies $\tilde{\pi}_h$ and $\tilde{\pi}_{h+1}$ -- i.e, utilizing \eqref{eq:standard-CI}.
\begin{equation}
\label{eq:standard-CI}
    \begin{aligned}
        &\E_{D(\pi_b)}[\supscript{\hatV_{\tilde{\pi}_h,o}}{h}(\supscript{x}{h})-\supscript{\hatV_{\tilde{\pi}_{h+1},p}}{h}(\supscript{x}{h})] \geq \supscript{\alpha}{h}_{\pi}\\
        &\geq \E_{D(\pi_b)}[\supscript{\hatV_{\tilde{\pi}_h,p}}{h}(\supscript{x}{h})-\supscript{\hatV_{\tilde{\pi}_{h+1},o}}{h}(\supscript{x}{h})]
    \end{aligned}
\end{equation}
For selective CIs, we use \eqref{eq:Q-TE-bound} to construct the lower bound and similarly construct the upper bound. Note that converting inequalities like \eqref{eq:Q-TE-bound} and \eqref{eq:standard-CI} into empirical bounds is straightforward since our training is from $D(\pi_b)$. \Cref{fig:chainbanditenv-CI} plots CIs for both selective and standard uncertainty propagation, when varying the evaluation policy. As expected, benefits over standard pessimism are larger when next-state distribution shift is smaller -- that is, when evaluation policy is closer to the behavioral policy. 

\begin{figure}[ht]
\begin{center}
\centerline{
\includegraphics[width=0.3\textwidth]{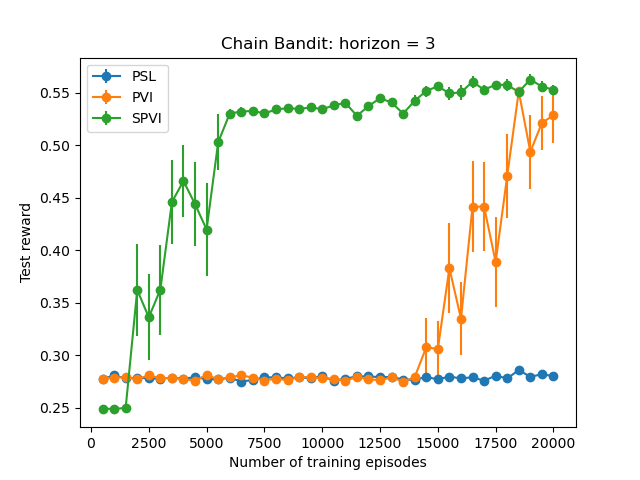}
}
\caption{Policy learning with a bad behavioral policy}
\label{fig:chainbandit-subopt}
\end{center}
\vskip -0.2in
\end{figure}

In \Cref{fig:chainbandit-subopt}, we plot the value of learnt policy from various algorithms as we vary the number of training episodes. In particular, we compare SPVI, PVI \citep{jin2021pessimism}, and pessimistic supervised learning (PSL). Here PSL refers to a pessimistic bandit policy optimization applied to each step without planning. The ChainBandit environment benefits from planning, so PSL performs poorly as expected. 

On all Chain Bandit simulations we tried, SPVI was by far the best-performing algorithm. The reason we considered the behavioral policy described earlier was that it was more disadvantageous for SPVI. In particular, since selective pessimism has an initial bias against policies that lead to significant shifts, we chose a highly sub-optimal behavioral policy (selecting $a_3$ with probability $0.8$). While this leads to a worse start for SPVI than PVI, eventually, SPVI outperforms the other algorithms. We also run simulations for CI construction and policy learning on the standard GridWorld (see \Cref{sec:additional-simulations}) -- since this is a very dynamic environment, both standard and selective propagation have similar performance.\footnote{All algorithm runs takes less than 2 mins on a MacBook Pro M2 16GB.}

\textbf{Conclusion:} We introduce selective propagation, a general approach to interpolate between CB and RL techniques -- achieving guarantees that adapt to instance hardness. Further developing this could impact real world problems (e.g., recommendation systems, mHealth, EdTech) that lie in between the two settings.

\bibliography{Brano, ref}

\newpage

\onecolumn

\title{Selective Uncertainty Propagation in Offline RL\\(Supplementary Material)}
\maketitle

\appendix

\section{Proof for Section \ref{sec:selective-uncertainty-propagation-theory}}

In this section we prove our main theoretical result, \Cref{thm:treatment-effect-CI}. For convenience, we restate it below.
\lemTE*

We now begin the proof of \Cref{thm:treatment-effect-CI}. We start by proving Lemma~\ref{lem:less-empirical-version-of-te-ci-thm}, which shows $\supscript{\tilde{\alpha}}{h}_{\pi}$ is close to our target estimand. Where $\supscript{\tilde{\alpha}}{h}_{\pi}$ is defined in \eqref{eq:less-emperical-te-estimator-at-step-h} and can be viewed as a less empirical version of $\supscript{\hat{\alpha}}{h}_{\pi}$ (our treatment effect estimator at step $h$). 
\begin{equation}
\label{eq:less-emperical-te-estimator-at-step-h}
    \supscript{\tilde{\alpha}}{h}_{\pi} = \supscript{\hat{\theta}}{h}_{\pi} + \E_{D(\pi_b)}\bigg[\int_{x'}\supscript{\hatV_{\pi}}{h+1}(x')\supscript{\hat{\Delta}}{h}(x'|\supscript{x}{h},\pi)\bigg]
\end{equation}

\begin{lemma}
\label{lem:less-empirical-version-of-te-ci-thm}
    Under the conditions of \Cref{thm:treatment-effect-CI}, we show the following bound holds with probability $1-\delta_{\text{in}}$:
    \begin{equation}
    \label{eq:less-emperical-close-to-estimand}
    \begin{aligned}
    &|\supscript{\tilde{\alpha}}{h}_{\pi} -  \supscript{\alpha}{h}_{\pi}|\leq \supscript{\tilde{L}_{\pi}}{h}. 
    \end{aligned}
    \end{equation}
    Where $\supscript{\tilde{L}_{\pi}}{h}$ is given by \eqref{eq:define-bound-less-empirical}.
    \begin{equation}
    \label{eq:define-bound-less-empirical}
        \begin{aligned}
        &\supscript{\tilde{L}_{\pi}}{h} = \supscript{\kappa}{h}_{\pi,\theta} + \supscript{V_{\max}}{h+1}\supscript{\kappa}{h}_{\pi,\Delta} + \E_{D(\pi_b)}\bigg[\int_{x'} |\E_{D(\pi_b)}[\supscript{\hat{\Delta}}{h}(x'|\supscript{x}{h},\pi)]| \cdot  \supscript{{\Gamma}_{\pi}}{h+1}(x')\bigg] 
        \end{aligned}
    \end{equation} 
\end{lemma}
\begin{proof}
    We start by simplifying $\supscript{\alpha}{h}_{\pi}$, defined in \eqref{eq:estimand-of-interest}.
    \begin{equation}
    \label{eq:simplifying-estimand}
        \begin{aligned}
            &\supscript{\alpha}{h}_{\pi} = \E_{D(\pi_b)}[\supscript{V}{h}_{\tilde{\pi}_h}(\supscript{x}{h}) - \supscript{V}{h}_{\tilde{\pi}_{h+1}}(\supscript{x}{h})]\\
            & \stackrel{(i)}{=} \E_{D(\pi_b)}\bigg[ \bigg(\supscript{R}{h}(x,\tilde{\pi}_h) + \int_{x'} \supscript{V_{\pi}}{h+1}(x')\supscript{P}{h}(x'|x,\tilde{\pi}_h) \bigg) - \bigg(\supscript{R}{h}(x,\tilde{\pi}_{h+1}) + \int_{x'} \supscript{V_{\pi}}{h+1}(x')\supscript{P}{h}(x'|x,\tilde{\pi}_{h+1})\bigg) \bigg] \\
            & \stackrel{(ii)}{=} \E_{D(\pi_b)}\bigg[ \bigg(\supscript{R}{h}(x,\tilde{\pi}_h) - \supscript{R}{h}(x,\tilde{\pi}_{h+1})  \bigg) + \bigg(\int_{x'} \supscript{V_{\pi}}{h+1}(x')\Big(\supscript{P}{h}(x'|x,\tilde{\pi}_h) - \supscript{P}{h}(x'|x,\tilde{\pi}_{h+1})\Big)\bigg)\bigg] \\
            & \stackrel{(iii)}{=} \supscript{\theta}{h}_{\pi} + \E_{D(\pi_b)}\bigg[ \int_{x'} \supscript{V_{\pi}}{h+1}(x')\supscript{\Delta}{h}(x'|x,\pi)\bigg]  
        \end{aligned}
    \end{equation}
    Here (i) follows from \eqref{eq:value-iteration-definition}, (ii) follows from re-arranging terms, and (iii) follows from \eqref{eq:immediate-TE-estimand} and \eqref{eq:shift-definition}.

    With probability $1-\delta_{\text{in}}$, we have the input guarantees of \Cref{sec:selective-uncertainty-propagation-theory} hold. Hence, utilizing these guarantees, we can bound the distance between $\supscript{\tilde{\alpha}}{h}_{\pi}$ and the target estimand $\supscript{\alpha}{h}_{\pi}$.
    \begin{equation}
        \begin{aligned}
            &|\supscript{\tilde{\alpha}}{h}_{\pi} -  \supscript{\alpha}{h}_{\pi}|\\
            &\stackrel{(i)}{=} \bigg|\bigg(\supscript{\hat{\theta}}{h}_{\pi} + \E_{D(\pi_b)}\bigg[\int_{x'}\supscript{\hatV_{\pi}}{h+1}(x')\supscript{\hat{\Delta}}{h}(x'|\supscript{x}{h},\pi)\bigg]\bigg) - \bigg(\supscript{\theta}{h}_{\pi} + \E_{D(\pi_b)}\bigg[ \int_{x'} \supscript{V_{\pi}}{h+1}(x')\supscript{\Delta}{h}(x'|x,\pi)\bigg] \bigg)\bigg|\\
            &\stackrel{(ii)}{\leq} \supscript{\kappa}{h}_{\pi,\theta} + \bigg|\E_{D(\pi_b)}\bigg[\int_{x'}\bigg(\supscript{\hatV_{\pi}}{h+1}(x')\supscript{\hat{\Delta}}{h}(x'|\supscript{x}{h},\pi) - \supscript{V_{\pi}}{h+1}(x')\supscript{{\Delta}}{h}(x'|\supscript{x}{h},\pi) \bigg)\bigg]\bigg|\\ 
            &\stackrel{(iii)}{\leq} \supscript{\kappa}{h}_{\pi,\theta} + \bigg|\E_{D(\pi_b)}\bigg[\int_{x'}\Big(\supscript{\hatV_{\pi}}{h+1}(x') - \supscript{V_{\pi}}{h+1}(x') \Big)\supscript{\hat{\Delta}}{h}(x'|\supscript{x}{h},\pi)\bigg]\bigg|\\ 
            &+ \bigg|\E_{D(\pi_b)}\bigg[\int_{x'}\supscript{V_{\pi}}{h+1}(x') \Big(\supscript{\hat{\Delta}}{h} - \supscript{{\Delta}}{h} \Big)(x'|\supscript{x}{h},\pi) \bigg) \bigg]\bigg|\\
            & \stackrel{(iv)}{\leq} \supscript{\kappa}{h}_{\pi,\theta} + \E_{D(\pi_b)}\bigg[\int_{x'} |\E_{D(\pi_b)}[\supscript{\hat{\Delta}}{h}(x'|\supscript{x}{h},\pi)]| \cdot  \supscript{{\Gamma}_{\pi}}{h+1}(x')\bigg] + \supscript{V_{\max}}{h+1}\supscript{\kappa}{h}_{\pi,\Delta} =: \supscript{\tilde{L}_{\pi}}{h} 
        \end{aligned}
    \end{equation}
    Here (i) follows from \eqref{eq:less-emperical-te-estimator-at-step-h} and \eqref{eq:simplifying-estimand}, (ii) follows from triangle inequality and the input guarantee, (iii) follows from triangle inequality, and finally (iv) follows from Cauchy-Schwarz inequality and the input guarantees.
\end{proof}
Having shown in Lemma~\ref{lem:less-empirical-version-of-te-ci-thm} that $\supscript{\tilde{\alpha}}{h}_{\pi}$ is close to our target estimand, we only need to show two more facts to complete the proof of \Cref{thm:treatment-effect-CI}: (i) we need to show $\supscript{\tilde{\alpha}}{h}_{\pi}$ is close to our treatment effect estimator $\supscript{\hat{\alpha}}{h}_{\pi}$ (defined in \eqref{eq:te-estimator-at-step-h}), and (ii) we need to show $\supscript{\tilde{L}_{\pi}}{h}$ is sufficiently smaller that $\supscript{{L}_{\pi}}{h}$. We show both these statements hold with high-probability.

We start with showing $\supscript{\tilde{\alpha}}{h}_{\pi}$ is close to $\supscript{\hat{\alpha}}{h}_{\pi}$ with high-probability. In particular, we have \eqref{eq:high-prob-alpha} holds with probability at least $1-\delta/2$.
\begin{equation}
\label{eq:high-prob-alpha}
    \begin{aligned}
        &|\supscript{\tilde{\alpha}}{h}_{\pi} -  \supscript{\hat{\alpha}}{h}_{\pi}|\\
        &=\bigg|\E_{D(\pi_b)}\bigg[\int_{x'}\supscript{\hatV_{\pi}}{h+1}(x')\supscript{\hat{\Delta}}{h}(x'|\supscript{x}{h},\pi)\bigg] - \frac{1}{T}\sum_{t=1}^{T}\int_{x'}\supscript{\hatV_{\pi}}{h+1}(x')\supscript{\hat{\Delta}}{h}(x'|\supscript{x}{h}_t,\pi) \bigg|\\ 
        &\leq 2\supscript{V_{\max}}{h+1}\sqrt{\frac{\ln(4/\delta)}{2T}}.
    \end{aligned}
\end{equation}
In \eqref{eq:high-prob-alpha}, the first equality follows from \eqref{eq:te-estimator-at-step-h} and \eqref{eq:less-emperical-te-estimator-at-step-h}. The first inequality follows from Hoeffding's inequality the fact that $\int_{x'}\supscript{\hatV_{\pi}}{h+1}(x')\supscript{\hat{\Delta}}{h}(x'|\supscript{x}{h}_t,\pi)\in [-\supscript{V_{\max}}{h+1},\supscript{V_{\max}}{h+1}]$.

From Hoeffding's inequality and the fact that \eqref{eq:sanity-check} holds, we also have \eqref{eq:high-prob-L} holds with probability at least $1-\delta/2$. 
\begin{equation}
\label{eq:high-prob-L}
    \begin{aligned}
        &\E_{D(\pi_b)}\bigg[\int_{x'} |\E_{D(\pi_b)}[\supscript{\hat{\Delta}}{h}(x'|\supscript{x}{h},\pi)]| \cdot  \supscript{{\Gamma}_{\pi}}{h+1}(x')\bigg] - \frac{1}{T}\sum_{t=1}^{T}\int_{x'} |\E_{D(\pi_b)}[\supscript{\hat{\Delta}}{h}(x'|\supscript{x}{h}_t,\pi)]| \cdot  \supscript{{\Gamma}_{\pi}}{h+1}(x')\\ 
        &\leq 4\supscript{V_{\max}}{h+1}\sqrt{\frac{\ln(4/\delta)}{2T}}.
    \end{aligned}
\end{equation}
Note that \eqref{eq:high-prob-L} implies $\supscript{\tilde{L}}{h}_{\pi} + 2\supscript{V_{\max}}{h+1}\sqrt{\frac{\ln(4/\delta)}{2T}}$ is no larger than $\supscript{L}{h}_{\pi}$.

Hence, with probability at least $1-\delta_{\text{in}}-\delta$, we have \eqref{eq:less-emperical-close-to-estimand}, \eqref{eq:high-prob-alpha}, and \eqref{eq:high-prob-L} hold. We now use these equation to show that \eqref{eq:final-bound-proof} holds.
\begin{equation}
\label{eq:final-bound-proof}
    \begin{aligned}
        &|\supscript{\alpha}{h}_{\pi}-\supscript{\hat{\alpha}}{h}_{\pi}|\\
        &\stackrel{(i)}{\leq} |\supscript{\alpha}{h}_{\pi}-\supscript{\tilde{\alpha}}{h}_{\pi}| + |\supscript{\tilde{\alpha}}{h}_{\pi}-\supscript{\hat{\alpha}}{h}_{\pi}|\\
        &\stackrel{(ii)}{\leq} \supscript{\tilde{L}_{\pi}}{h} + 2\supscript{V_{\max}}{h+1}\sqrt{\frac{\ln(4/\delta)}{2T}}\\ 
        &\stackrel{(iii)}{\leq} \supscript{{L}_{\pi}}{h}.
    \end{aligned}
\end{equation}
Here (i) follows from triangle inequality; (ii) follows from \eqref{eq:less-emperical-close-to-estimand} and \eqref{eq:high-prob-alpha}; and (iii) follows from \eqref{eq:high-prob-L}. Therefore, we have shown that \eqref{eq:final-bound-proof} holds with probability at least $1-\delta_{\text{in}}-\delta$. This completes the proof of \Cref{thm:treatment-effect-CI}.

\section{SPVI Pseudocode}
\label{sec:spvi-pseudocode}
\begin{algorithm}[H]%
\caption{Selectively Pessimistic Value Iteration}\label{alg:SPVI}
\begin{algorithmic}
\State Inputs: Estimated reward model $\hat{R}$, estimated transition model $\hat{P}$, induced shift model $\hat{\Delta}$, and estimation uncertainty bonuses $b$.
\State Initialize $\supscript{\hat{V}}{H+1}, \supscript{\hat{V}_p}{H+1}, \supscript{\hat{V}_o}{H+1} \equiv 0$.
\For {$h=H \mbox{ to } 1$}
\State For all $(x,a)\in\calX\times\calA$,
\Comment{Estimate Q value.}
\begin{equation}
\label{eq:empirical-Q-value}
    \begin{aligned}
        \supscript{\hatQ}{h}(x,a) \leftarrow \supscript{\hatR}{h}(x,a) + \int_{x'} \supscript{\hatV}{h+1}(x') \supscript{\hat{P}}{h}(x'|x,a)
    \end{aligned}
\end{equation}
\State For all $(x,a)\in\calX\times\calA$,
\Comment{Selective pessimism.}
\begin{equation}
\label{eq:selectively-pessimistic-Q-value}
    \begin{aligned}
        &\supscript{\hatQ_{\text{sp}}}{h}(x,a) \leftarrow \supscript{\hatQ}{h}(x,a)-\supscript{b}{h}(x,a)- \int_{x'} |\supscript{\hat{\Delta}}{h}(x'|x,a)| \cdot (\supscript{\hatV_{o}}{h+1}(x') - \supscript{\hatV_{p}}{h+1}(x')),\\
        &\supscript{{\pi}}{h}(x) \leftarrow \argmax_a \supscript{\hatQ_{\text{sp}}}{h}(x,a)
    \end{aligned}
\end{equation}
\State For all $x\in\calX$,
\Comment{Estimate pessimistic value.}
\begin{equation}
\label{eq:pessimistic-value-estimate}
    \begin{aligned}
        &\supscript{\hatV_{\text{p}}}{h}(x) \leftarrow \max\Big(0,\;\supscript{\hatR}{h}(x,\supscript{\pi}{h}(x))+ \int_{x'} \supscript{\hatV_p}{h+1}(x') \supscript{\hat{P}}{h}(x'|x,a)-\supscript{b}{h}(x,a)\Big)
    \end{aligned}
\end{equation}
\State For all $x\in\calX$,
\Comment{Estimate optimistic value.}
\begin{equation}
\label{eq:optimistic-value-estimate}
    \begin{aligned}
        &\supscript{\hatV_{o}}{h}(x) \leftarrow \min\Big( \supscript{V_{\max}}{h},\; \supscript{\hatR}{h}(x,\supscript{\pi}{h}(x))+ \int_{x'} \supscript{\hatV_o}{h+1}(x') \supscript{\hat{P}}{h}(x'|x,a)+\supscript{b}{h}(x,a)\Big)
    \end{aligned}
\end{equation}
\State For all $x\in\calX$,
\Comment{Estimate value.}
\begin{equation}
\label{eq:bounded-sp-value-estimate}
    \begin{aligned}
        &\supscript{\hatV}{h}(x)\leftarrow \min\Big(\supscript{\hatV_{o}}{h}(x),\max\Big(\supscript{\hatV_{p}}{h}(x),\supscript{\hatQ}{h}(x,\supscript{\pi}{h}(x))\Big)\Big)
    \end{aligned}
\end{equation}
\EndFor
\end{algorithmic}
\end{algorithm}

\section{Additional Simulation}
\label{sec:additional-simulations}

Similar to \Cref{sec:simulation}, we now run simulations for the standard GridWorld environment (with width $8$ and height $3$). Here states are discrete points on a bounded two-dimensional grid. The agent’s starting state is sampled uniformly at random from the grid, and the agent should learn to reach a specified goal state (which is an absorbing state). Upon transitioning to the goal state, the agent receives a reward of one and receives a reward of zero otherwise. The agent has 4 actions (left, right, up, and down). These actions make the agent move one step in that direction if possible. If the agent is at the boundary of the grid and can’t move in the direction selected, the agent continues to stay in the same state. Since we make the goal state an absorbing state, all actions at this state lead to the agent continuing to stay in this state. We set the starting state for the GridWorld as (1,1), the terminal state as (2,2), and the horizon as 3. For all states/steps, our behavioral policy samples actions (left, right, up, and down) with the probabilities $(0.20, 0.10, 0.50, 0.20)$. Other choices for GridWorld environment parameters and behavioral policy appear to generate similar plots. All our plots in this section are averaged over five simulation runs.

\begin{figure}[H]
\begin{center}
\centerline{
\includegraphics[width=0.6\textwidth]{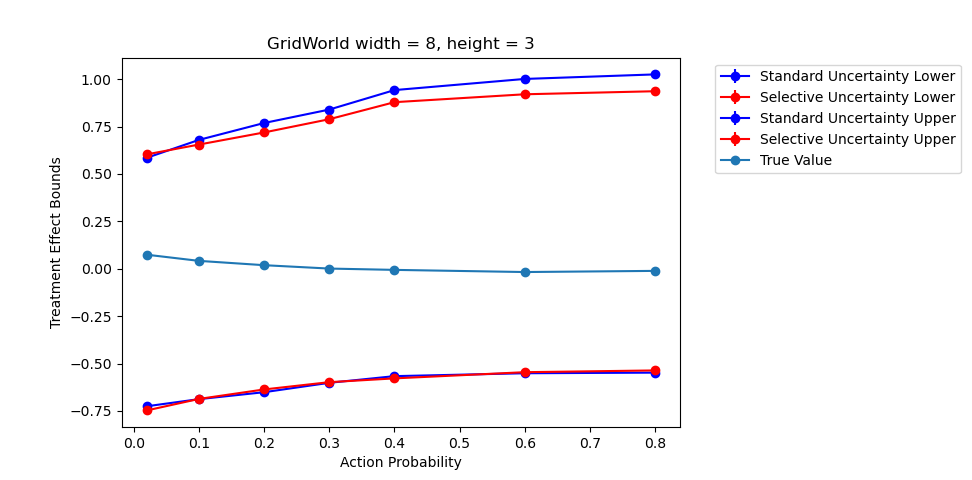}
}
\caption{Plotting $\supscript{\alpha}{2}_{\pi}$ with varying the evaluation policy $\pi$. Here $\lambda$ (evaluation policy probability of taking the down action) corresponds to the X-axis.}
\label{fig:gridworldenv-CI}
\end{center}
\vskip -0.25in
\end{figure} 

We constructed intervals for $\supscript{\alpha}{2}_{\pi}$, the treatment effect at step $2$, using both selective and standard uncertainty propagation. In \Cref{fig:gridworldenv-CI}, we plot the CIs for $\supscript{\alpha}{2}_{\pi}$ as we vary the evaluation policy. With $\lambda$ parameterizing the evaluation policies. For all states/steps, the evaluation policy corresponding to $\lambda$ samples actions (left, right, up, and down) with the probabilities $(0.25, 0.20, 0.55-\lambda, \lambda)$. Our CIs are constructed from sampling a dataset of $2000$ episodes. In \Cref{fig:gridworldenv-CI}, we see that both selective and standard uncertainty propagation have a similar performance -- this is understandable because GridWorld is a dynamic environment (each action has a different next-state distribution), hence estimate/uncertainty propagation is less avoidable here for valid CI construction.

\begin{figure}[H]
\begin{center}
\centerline{
\includegraphics[width=0.5\textwidth]{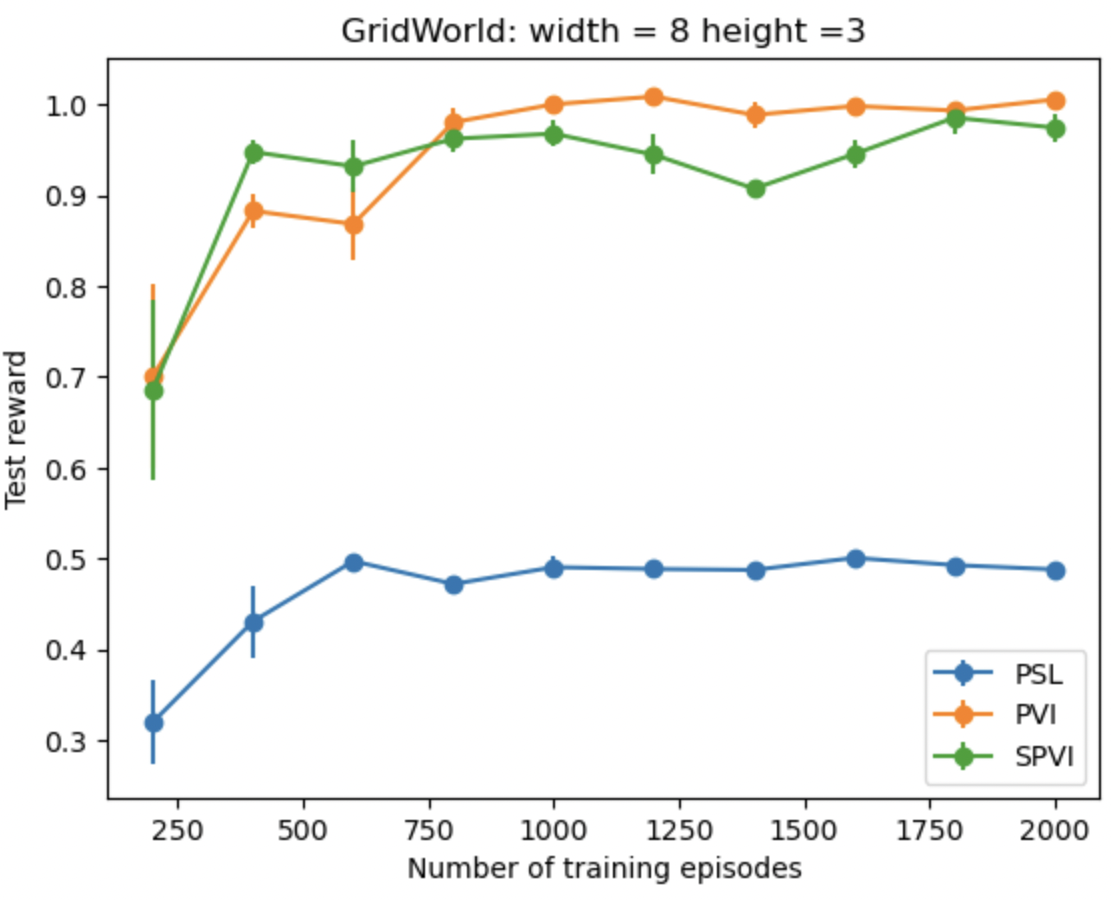}
}
\caption{Learning Experiment on GridWorld}
\label{fig:gridworldenv-Opt}
\end{center}
\vskip -0.25in
\end{figure} 

We also compare the policy learning algorithms on the same GridWorld environment, with the same behavioral policy.  In \Cref{fig:gridworldenv-Opt}, we plot the value of policy learnt by SPVI, PVI, and PSL -- as we vary the number of training episodes. We see PSL still performs terribly since planning is necessary. However, since GridWorld is a dynamic environment, PVI and SPVI have a similar performance.

\end{document}